\documentclass[final,3p,times]{elsarticle}
\usepackage{graphicx}
\usepackage{amssymb}
\usepackage{framed}
\usepackage{lineno}

\journal{arXiv}

\begin{document}

\begin{frontmatter}


\title{Efficient Calculation of Bigram Frequencies in a Corpus of Short Texts}


\author{Melvyn Drag}
\author{Gauthaman Vasudevan}
\address{Avlino, Inc. Holdmel, NJ}

\begin{abstract}
We show that an efficient and popular method for calculating bigram frequencies is unsuitable for bodies of short texts and offer a simple alternative. Our method has the same computational complexity as the old method and offers an exact count instead of an approximation.
\end{abstract}

\begin{keyword}
NLP \sep Machine Learning
\end{keyword}

\end{frontmatter}

\section{Acknowledgements}
\label{S:-1}
This short note is the result of a brief conversation between the authors and Joel Nothman. We came across a potential problem, he gave a sketch of a fix, and we worked out the details of a solution.
\section{Calculating Bigram Frequecies}
\label{S:0}
A common task in natural language processing is to find the most frequently occurring word pairs in a text(s) in the expectation that these pairs will shed some light on the main ideas of the text, or offer insight into the structure of the language. One might be interested in pairings of adjacent words, but in some cases one is also interested in pairs of words in some small neighborhood. The neighborhood is usually refered to as a window, and to illustrate the concept consider the following text and bigram set:
\begin{framed}
\noindent Text: ``I like kitties and doggies''
\newline Window: 2
\newline Bigrams: \{(I like), (like kitties), (kitties and), (and doggies)\}
\end{framed}
and this one:
\begin{framed}
\noindent Text: ``I like kitties and doggies''
\newline Window: 4
\newline Bigrams: \{(I like), (I kitties), (I and), (like kitties), (like and), (like doggies), (kitties and), (kitties doggies), (and doggies)\}.
\end{framed}

\section{The Popular Approximation}
\label{S:1}
Bigram frequencies are often calculated using the approximation
\begin{equation}\label{eq:1}
freq(*, word) = freq(word, *) = freq(word)
\end{equation}
In a much cited paper, Church and Hanks~\cite{Church:1990} use `$=$' in place of `$\approx$' because the approximation is so good. Indeed, this approximation will only cause errors for the very few words which occur near the beginning or the end of the text. Take for example the text appearing above - the bigram (doggies, *) does not occur once, but the approximation says it does. 

An efficient method for computing the contingency matrix for a bigram (word1, word2) is suggested by the approximation. Store $freq(w1, w2)$ for all bigrams $(w1, w2)$ and the frequencies of all words. Then,
\begin{itemize}
\item$freq(word1, word2)$ is known, \item$freq(\sim word1, word2) \approx freq(word2)$ - $freq(word1, word2)$, 
\item $freq(word1, \sim word2) \approx freq(word1)$ - $freq(word1, word2)$, 
\item and $freq(\sim word1, \sim word2)$ is easily computed.
\end{itemize}

The statistical importance of miscalculations due to this method diminishes as our text grows larger and larger. Interest is growing in the analysis of small texts, however, and a means of computing bigrams for this type of corpus must be employed. This approximation is implemented in popular NLP libraries and can be seen in many tutorials across the internet. People who use this code, or write their own software, must know when it is appropriate.
\section{An Alternative Method}
\label{S:2}
We propose an alternative. As before, store the frequencies of words and the frequencies of bigrams, but this time store two additional maps called \textbf{too\_far\_left} and \textbf{too\_far\_right}, of the form \{word : list of offending indices of word\}. The offending indices are those that are either too far to the left or too far to the right for approximation~(\ref{eq:1}) to hold. All four of these structures are built during the construction of a bigram finder, and do not cripple performance when computing statistical measures since maps are queried in $O(1)$ time.

As an example of the contents of the new maps, in ``Dogs are better than cats", \textbf{too\_far\_left[`dog'] = [0]} for all windows. In ``eight mice eat eight cheese sticks'' with window 5,  \textbf{too\_far\_left[`eight'] = [0,3]}.
For ease of computation the indices stored in \textbf{too\_far\_right} are transformed before storage using:
\begin{equation}\label{eq:2}
\widehat{idx} = length - idx - 1 = g(idx)
\end{equation}
where $length$ is the length of the small piece of text being analyzed. Then, \textbf{too\_far\_right[`cats'] = [$g(4)= idx$] = [$0 = \widehat{idx}$]}.

Now, to compute the exact  number of occurrences of a bigram we do the computation:
\begin{equation}\label{eq:3}
freq(*, word) = (w-1)*wordfd[word] - \sum\limits_{i=1}^{N}(w-tfl[word][i] - 1)
\end{equation}
where $w$ is the window size being searched for bigrams, $wfd$ is a frequency distribution of all words in the corpus, $tfl$ is the map \textbf{too\_far\_left} and $N$ is the number of occurrences of the $word$ in a position too far left.The computation of $freq(word, *)$ can now be performed in the same way by simply substituting $tfl$ with $tfr$ thanks to transformation $g$, which reverses the indexing.
\nocite{*}
\bibliographystyle{amsplain}
\bibliography{bigram_bib.bib}

\providecommand{\bysame}{\leavevmode\hbox to3em{\hrulefill}\thinspace}
\providecommand{\MR}{\relax\ifhmode\unskip\space\fi MR }
\providecommand{\MRhref}[2]{%
  \href{http://www.ams.org/mathscinet-getitem?mr=#1}{#2}
}
\providecommand{\href}[2]{#2}
\begin{thebibliography}{1}

\bibitem{Bird:2009}
S.~Bird, E.~Klein, and E.~Loper, \emph{{N}atural {L}anguage {P}rocessing {W}ith
  {P}ython}, 1st ed., O'Reilly Media, Inc., 2009.

\bibitem{Church:1990}
K.~W. Church and P.~Hanks, \emph{{W}ord {Association} {N}orms, {M}utual
  {I}nformation, {A}nd {L}exicography}, {C}omputational {L}inguistics
  \textbf{16} (1990), no.~1, 22--29.

\end{thebibliography}
\end{document}